\documentclass[conference]{IEEEtran}
\IEEEoverridecommandlockouts
\usepackage{cite}
\usepackage{amsmath,amssymb,amsfonts}
\usepackage{algorithmic}
\usepackage{graphicx}
\usepackage{textcomp}
\usepackage{xcolor}
\def\BibTeX{{\rm B\kern-.05em{\sc i\kern-.025em b}\kern-.08em
    T\kern-.1667em\lower.7ex\hbox{E}\kern-.125emX}}
\begin{document}

\title{AttenFace: A Real Time Attendance System Using Face Recognition
}

\author{\IEEEauthorblockN{Ashwin Rao}
\IEEEauthorblockA{
\textit{International Institute of Information Technology, Hyderabad}\\
ashwin.rao@students.iiit.ac.in}
}

\maketitle

\begin{abstract}
The current approach to marking attendance in colleges is tedious and time consuming. I propose AttenFace, a standalone system to analyze, track and grant attendance in real time using face recognition. Using snapshots of class from live camera feed, the system identifies students and marks them as present in a class based on their presence in multiple snapshots taken throughout the class duration. Face recognition for each class is performed independently and in parallel, ensuring that the system scales with number of concurrent classes. Further, the separation of the face recognition server from the back-end server for attendance calculation allows the face recognition module to be integrated with existing attendance tracking software like Moodle. The face recognition algorithm runs at 10 minute intervals on classroom snapshots, significantly reducing computation compared to direct processing of live camera feed. This method also provides students the flexibility to leave class for a short duration (such as for a phone call) without losing attendance for that class. Attendance is granted to a student if he remains in class for a number of snapshots above a certain threshold. The system is fully automatic and requires no professor intervention or any form of manual attendance or even camera set-up, since the back-end directly interfaces with in-class cameras. AttenFace is a first-of-its-kind one-stop solution for face-recognition-enabled attendance in educational institutions that prevents proxy, handling all aspects from students checking attendance to professors deciding their own attendance policy, to college administration enforcing default attendance rules.

\end{abstract}


\section{Introduction}
Attendance is a mandatory part of every class in colleges. Often, there is a minimum attendance requirement for courses taken by students. The simplest methods of taking attendance include roll-call or manually signing on a document. These methods are tedious and waste time and do not take advantage of technology in any way. Attendance will have to be manually entered from the attendance sheet into the database. Further, proxy attendance is easy, wherein a student gives attendance to another student, either by forging his signature or calling out his name during roll-call. A current solution to make attendance easier, and gaining popularity, is biometric attendance. The biometric machine can be connected to the database and update attendance automatically. The problem of proxy is solved due to the uniqueness of thumb prints. However, this method still wastes time due to the students needing to queue up for biometric attendance. Passing the biometric machine around in class solves this issue, but can be disturbing. However, all the aforementioned methods of taking attendance still share a common problem: there is no way to ensure that students sit in class through its entire duration. A student can leave class immediately after attendance, or enter class just before attendance.

AttenFace's novel snapshot technique of face recognition (described in Sec. \ref{sec:sys-arch}) solves all these problems, including the issue of proxy attendance which is common in colleges. Students no longer have to manually give attendance, as their presence is automatically recognized by a camera. Since the camera is recording at all times, it is easy to capture how long a student remains in class. Final attendance can be given only if the student remained in class above a certain threshold of time, which can be decided by the professor teaching the class. Further, the system includes an easy-to-use portal for students to check their attendance for any class and course, and for professors to override default attendance rules for a particular class or student if necessary.

\section{Related Work}
The proposed attendance system requires three major technologies to identify students: 1) object detection and localization, to identify which objects in the classroom are students, and where they are, 2) face detection, to identify which object is a face, and 3) face recognition, to map detected faces to corresponding students. There is continuous research going on in these areas. YOLO \cite{b1} is a real-time object detection algorithm which is being continuously improved upon since its first introduction in 2016. Haar Casading \cite{b2} identifies faces in a real-time video feed. FaceNet \cite{b3} is a notable face-recognition technique, which uses a deep convolutional neural network with 22 layers trained via a triplet loss function to directly output a 128-dimensional embedding. VGGFace2 \cite{b4} is a dataset for training face recognition models taking into account pose and age. 

At the heart of the proposed system is the face recognition algorithm. This problem has numerous approaches \cite{b5}. Principal Component Analysis (PCA) extracts principal features to recognize faces. Template matching involves comparing and matching patterns to recognize faces, usually through a neural network. Taking into account pose variations and liveliness detection can help make the system more robust. Liveliness detection is the problem of differentiating between the face of a live person and a photograph \cite{b6}.

In recent years, a number of face recognition based attendance systems have been proposed. In \cite{b7}, face recognition along with Radio Frequency Identification (RFID) was used to detect authorized students and count the number of times a student enters and leaves the classroom. In \cite{b8}, students were recognized through iris biometrics. The system automatically took attendance by capturing an image of the eye of each student and searching for a match in the database. In \cite{b9}, the Eigenface and Fisherface face recognition algorithms were compared and used in a real-time attendance system. Eigenface was found to perform better, with an accuracy of 70-90\%. In \cite{b10}, authors used Discrete Wavelet Transform (DWT) and Discrete Cosine Transform (DCT) to extract features of students' faces, followed by the Radial Basis Function (RBF) to recognize them. Their attendance system had an accuracy of about 82\%. In \cite{b11}, authors considered various real-time scenarios such as lighting and pose of students. A 3D approach to recognize faces for attendance was put forward in \cite{b12}.

There are a number of attendance systems available on the market which use face recognition for identification. Most of these, however, are time and attendance systems, which are used for one-time close-up identification of a single person at a time. For example, Truein \cite{b13} is a touchless face recognition system used to manage employee attendance in the workplace. iFace \cite{b14} provides face recognition capabilities through a mobile app, useful in work-from-home scenarios. There is currently no product in the market aimed at real-time face recognition for attendance capture in schools and colleges that provides a single unified portal to track attendance and modify attendance policy, which can also integrate with existing attendance management systems. The underlying technology of face recognition, however, is the same, and can be adapted to suit the proposed system.

\section{System architecture}
\label{sec:sys-arch}

The proposed system uses face recognition to automatically handle attendance of students. Taking the actual face recognition algorithm as a black box, the system decides attendance in the following manner:
\begin{itemize}
\item Recording begins at the start of the class. The start time of the class and the room number are available via the database.
\item A snapshot of the class is taken every 10 minutes. Using the snapshot, the face recognition algorithm recognizes students and marks them as present in that 10 minute block of time.
\item A student will be marked present for a class if he is present in at least 'n' snapshots. This threshold can be decided by the professor. This allows a student to leave the class in between in case of an emergency without losing attendance.
\end{itemize}

AttenFace's snapshot model provides a method to continuously track attendance throughout the class duration while avoiding the computationally expensive process of face recognition on live video. Simultaneously, it ensures that students must remain in class for a minimum amount of time to receive attendance, solving the issue of students leaving class after manual attendance. This makes the system equally efficient but more robust than existing solutions for face recognition, which do not run continuously during the whole class but rather involve a single verification before or after class.

\subsection{Requirements}\label{AA}
The requirements of the system are as follows:
\begin{enumerate}
\item Functional requirements
\begin{itemize}
    \item A login portal connected to the institute login, to be used by students, professors and administrators.
    \item A dashboard for students and professors to view attendance details for any class or course taken by them.
    \item An easy way for professors to make minor changes to attendance policy for a class or the entire course directly from the dashboard without needing to go through the administration.
    \item An option on the dashboard for the administration to manually override faulty attendance results.
\end{itemize}

\item Non-functional requirements
\begin{itemize}
    \item The system should be able to access required student information, course information and class information from the institute database.
    \item Portal to view attendance should be cross-platform with emphasis on mobile friendliness.
    \item The face recognition algorithm should be able to recognize faces in real-time, without much computational overhead.
    \item Multiple instances of the face recognition algorithm should be able to run in parallel, since there can be multiple classes going on at any given time.
\end{itemize}

\end{enumerate}

\subsection{Use Cases}\label{AA}

\begin{figure}[htbp]
\centering
\includegraphics[width=0.5\textwidth]{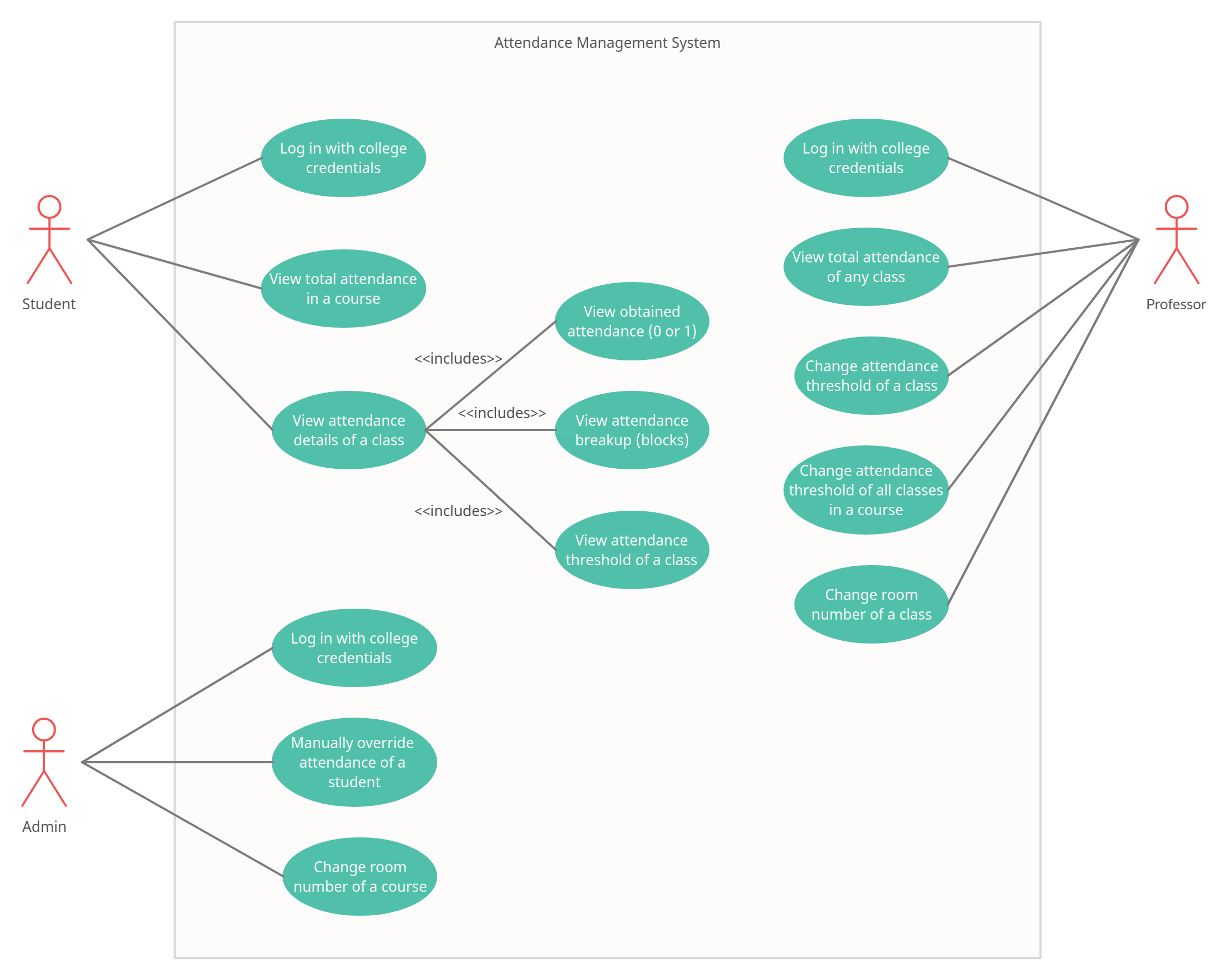}
\caption{UML use case diagram.}
\label{fig}
\end{figure}

\begin{enumerate}
\item A user (student, professor or admin) should be able to log in using their college credentials.
\item A student should be able to view his attendance for any class of any course he has taken.
\item A student should be able to view his attendance (0 or 1) for a class as soon as it ends.
\item A student should be able to see how many "blocks" in a class he attended (as explained in the previous section) and whether this is above the threshold set by the professor, hence justifying the attendance obtained for that class.
\item A student should be able to view the total number of classes he has attended in a course so far, and the number of classes he is allowed to miss before his total attendance for that course drops below the course's requirements.
\item A professor should be able to view the total attendance of a class as soon as it ends.
\item A professor should be able to change the threshold determining for how many blocks a student must be present in the class to get attendance. This can be changed for a specific class or for all classes in the course. This gives the professor the power to make attendance lenient or optional on a specific day, without having to go through the college administration.
\item A professor should be able to change the room number of the class before it starts, in case of any sudden events to ensure that attendance will still be taken by activating the camera in the new room.
\item An administrator should be able to directly change the attendance of a student in a particular class if required, overriding the automated attendance.
\item An administrator should be able to change the room number of the entire course, in case of any clashes with other classes or events.
\end{enumerate}

\subsection{System Architecture}\label{AA}

\begin{figure}[htbp]
\centering
\includegraphics[width=0.5\textwidth]{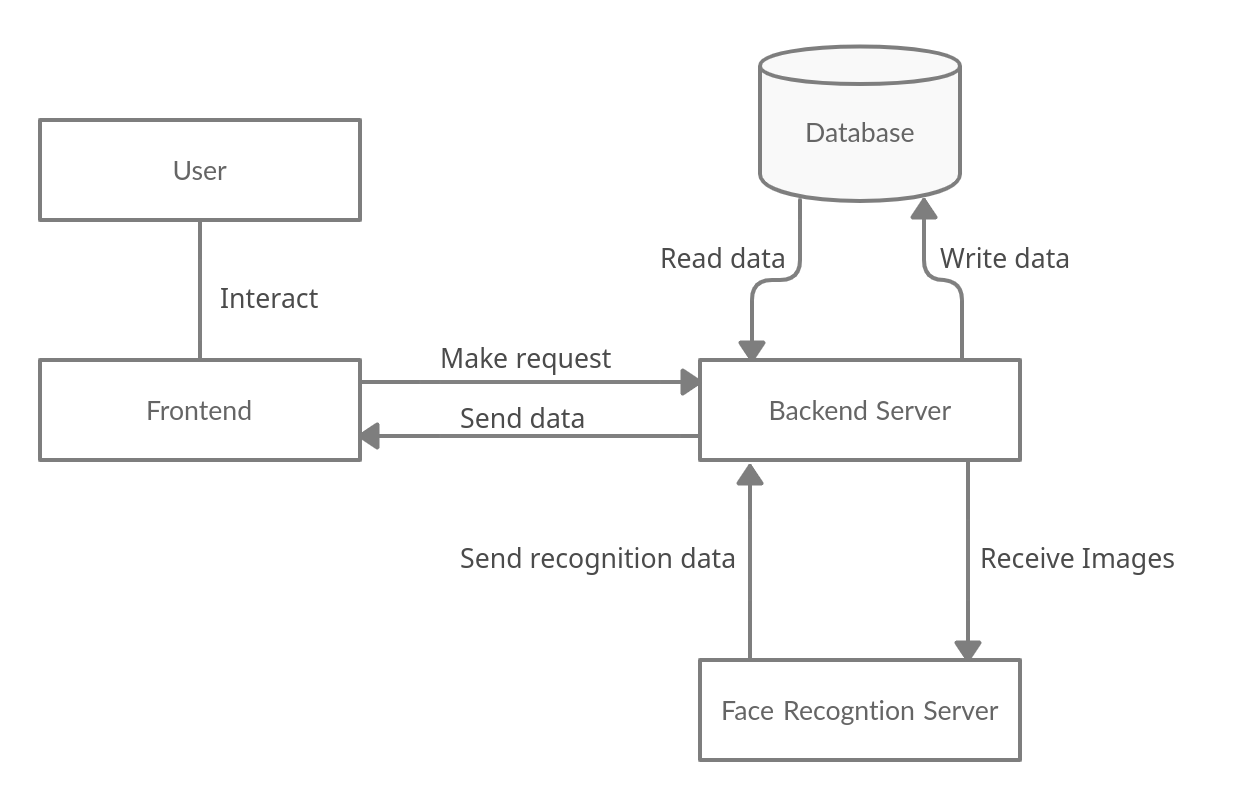}
\caption{System architecture.}
\label{fig}
\end{figure}

The system can be divided into the following modules:
\begin{enumerate}
\item Front-End (mobile and web application): All users interact with the system through the frontend. The following information will be displayed for a student: a) total attendance received so far in a particular course, b) attendance received in any class of any course he has registered for, c) the total "blocks" attended in any class to justify the attendance for that class, and d) the threshold to determine whether attendance is obtained or not in a particular class. The following information will be displayed for a professor: a) total attendance for any class in any course taught by him, and b) the threshold to determine attendance (which he can edit).

\begin{figure}[htbp]
\centering
\includegraphics[width=0.5\textwidth]{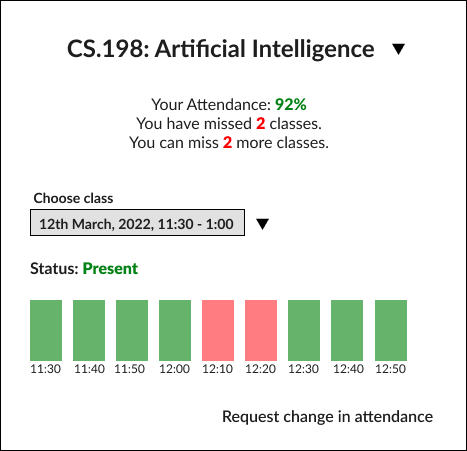}
\caption{Sample interface for students.}
\label{fig:student-gui}
\end{figure}

\begin{figure}[htbp]
\centering
\includegraphics[width=0.5\textwidth]{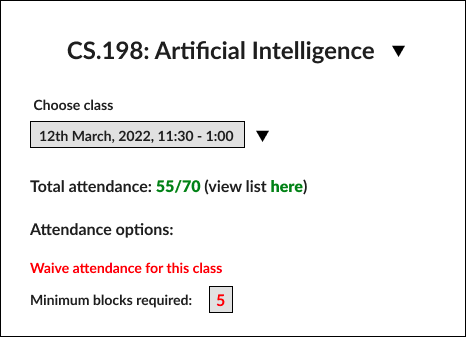}
\caption{Sample interface for professors.}
\label{fig:prof-gui}
\end{figure}

\item Back-End: The back-end handles all interactions with the database including accessing student images for the face recognition algorithm, displaying all relevant data on the front-end, and standard CRUD operations. It also performs the following calculations: a) total attendance of a student in a course given his attendance in all classes of the course so far, and b) attendance of a student in a class of a course given his attendance in each block of time in that class. The back-end acts as a bridge to send data to the face recognition server, which cannot access the database directly.

\item Face Recognition Model Server: The computationally heavy operations for face recognition will be performed here. Each ongoing class will have its own thread of computation associated with it, which obtains live feed directly from the concerned camera, and communicates with the back-end server for attendance calculation. Each thread of the face recognition server receives the following data from the back-end: a) Images of all students attending the class, b) start time of the class, c) end time of the class, and d) the camera ID to activate and obtain live feed from. The interaction diagram of the face recognition server with the back-end and camera is shown in Fig. 6. The direct communication between the back-end and camera is an integral component of making the system standalone and fully automatic compared to existing solutions.

\item Database: It stores information regarding students (such as images for recognition and the courses they have registered for), courses (such as room number and the corresponding camera ID), and professors (such as courses they are teaching). Fig. 8 shows a simplified schema of the database.
\end{enumerate}

\begin{figure}[htbp]
\centering
\includegraphics[width=0.5\textwidth]{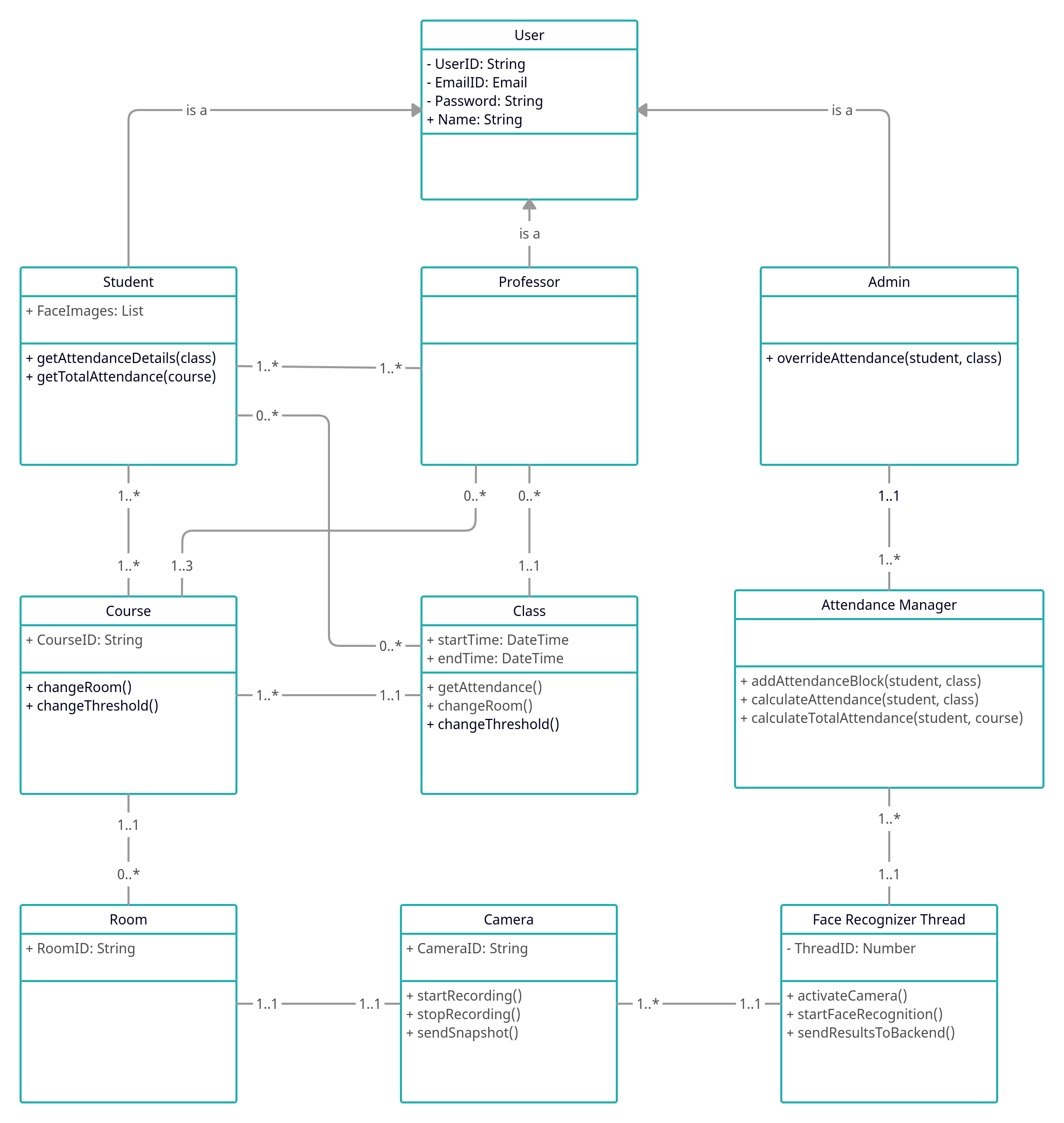}
\caption{UML class diagram.}
\label{fig}
\end{figure}

\begin{figure}[htbp]
\centering
\includegraphics[width=0.5\textwidth]{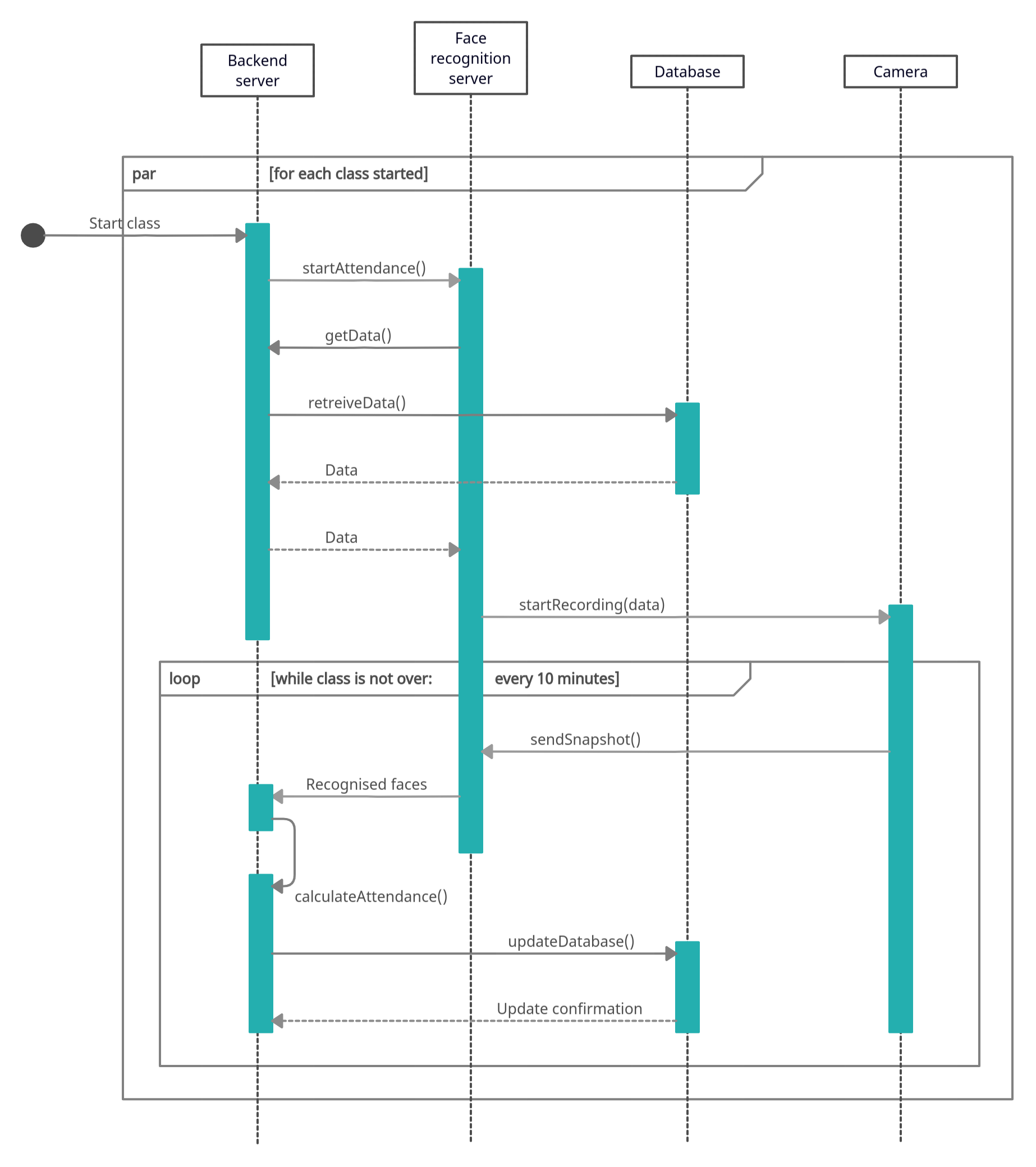}
\caption{UML 2.0 sequence diagram showing interactions between the back-end server, database and face recognition server.}
\label{fig}
\end{figure}

\begin{figure}[htbp]
\centering
\includegraphics[width=0.5\textwidth]{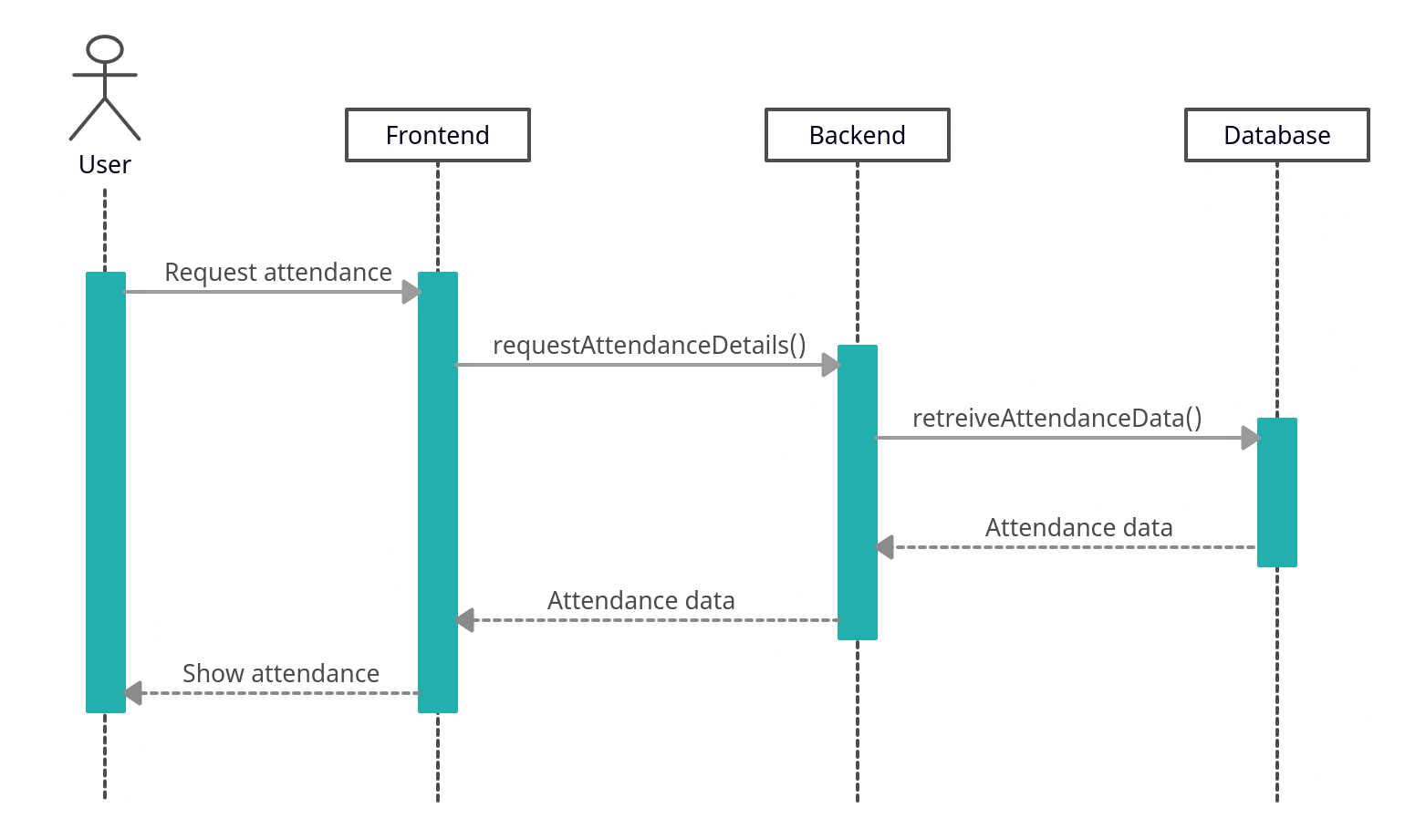}
\caption{UML sequence diagram showing interaction between user and front-end.}
\label{fig}
\end{figure}

% \newpage

\begin{figure}[htbp]
\centering
\includegraphics[width=0.5\textwidth]{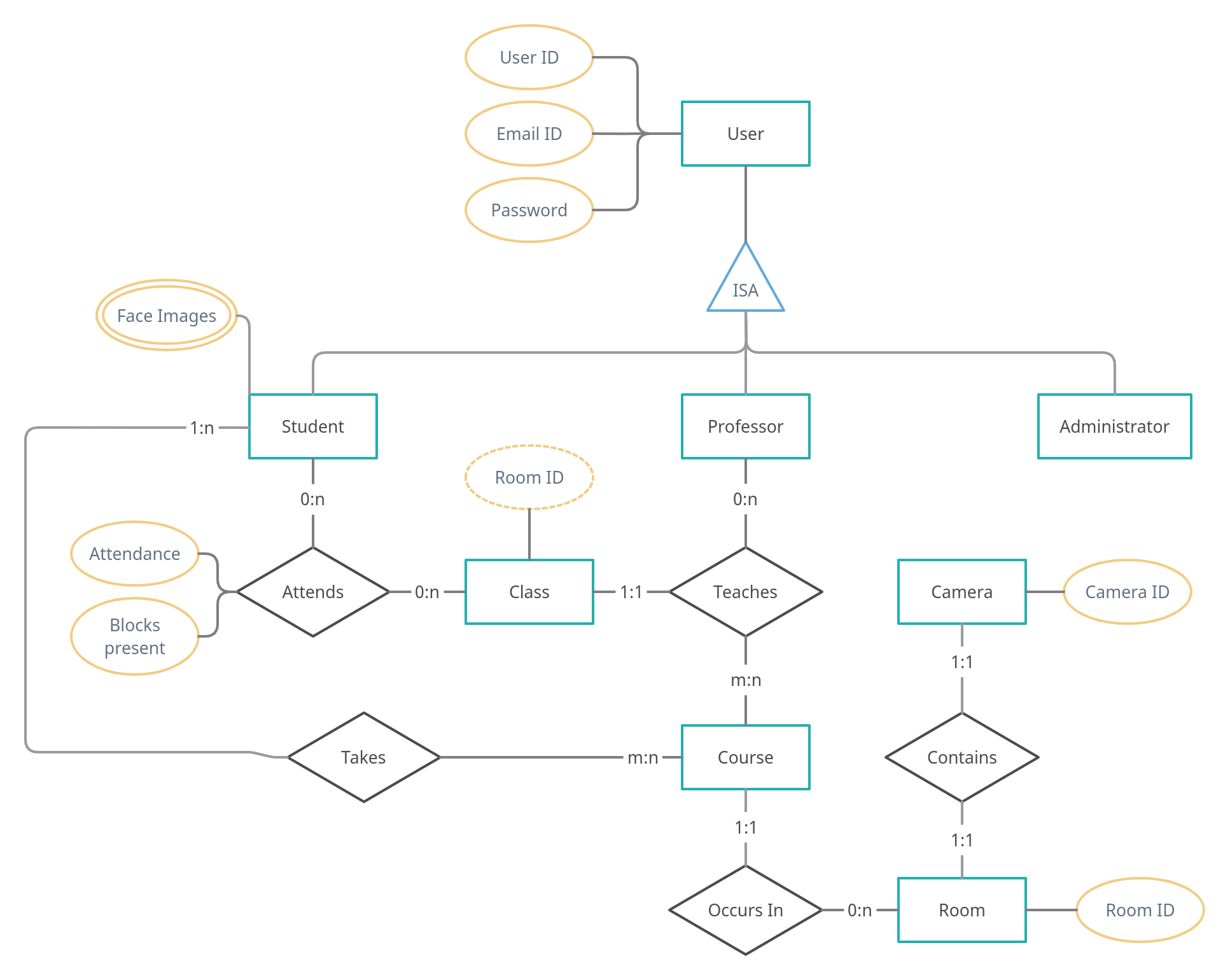}
\caption{Simplified database design.}
\label{fig}
\end{figure}

\subsection{University Classroom: A Practical Example}
A walkthrough of using AttenFace for a camera-enabled classroom in a university would be as follows:
\begin{itemize}
    \item The system establishes a connection with the camera 5 minutes before class starts.
    \item The professor, if he wishes to do so, logs into the portal and change attendance requirements for the current class (see Fig. 4).
    \item Starting from the beginning of class, and until class ends, a snapshot of the classroom is sent to the back-end and subsequently, the face recognition server, every 10 minutes. Students are identified and their presence is marked in that 10-minute block of time.
    \item After class ends, students can log in to the portal and immediately view their attendance status for that class (see Fig 3).
\end{itemize}

\subsection{Extensibility and Ease of Integration}\label{AA}
The proposed system is modular. The face recognition server, in particular, is a standalone module, which plays no role in the actual attendance policy of the professor or institute. Given pictures of students as input, all recognition-based calculations are done within the face recognition module and results for each student (whether present or not for that block of time) are returned to the back-end server, which handles attendance calculation using this data. Due to its modular nature, the system can be easily integrated into existing college portals. For example, integrating the proposed real-time attendance system with moodle is straightforward. The front-end, institute login and interaction with the college database are already handled by moodle. The only part of the system which needs to be integrated is a custom back-end script which interacts with the face recognition server and performs the desired calculations, and the face recognition server itself. The attendance data can then be made available to moodle to display on the frontend.

\section{Conclusion and future work}
This paper proposes a new method to analyze and grant attendance in real time using face recognition.
Attendance in each class is determined automatically with no human effort. The system ensures that a student must stay in class for at least a certain amount of time to be marked present, and at the same time grants a certain amount of leniency in the attendance calculation, as decided by the professor. There is always scope for improvement. Face recognition techniques are not completely accurate, and the system may sometimes be unable to identify students, or recognize them incorrectly. External factors such as classroom lighting and position of students faces may have an effect on the accuracy of the face recognition algorithm. As new research leads to better performing face recognition algorithms which are more robust and adaptable to varying situations, the proposed system benefits.

\end{document}